\pdfoutput=1

\documentclass[11pt]{article}

\usepackage[final]{acl}

\usepackage{times}
\usepackage{latexsym}

\usepackage[T1]{fontenc}

\usepackage[utf8]{inputenc}

\usepackage{microtype}
\usepackage{inconsolata}

\usepackage{textcomp}
\usepackage{amssymb}
\usepackage{adjustbox}
\usepackage{amsmath}
\usepackage{amssymb}
\usepackage{graphicx}
\usepackage{tabularx}
\usepackage{algorithm}
\usepackage{dsfont}
\usepackage{mathrsfs}
\usepackage{multirow}
\usepackage{tabularx}
\usepackage{color, colortbl}
\usepackage{caption}
\usepackage{booktabs}
\usepackage{subcaption}
\usepackage{mwe}
\usepackage{soul}
\usepackage{xurl}
\usepackage{amsfonts}
\usepackage{enumitem}
\usepackage{bm}
\usepackage{euflag}
\usepackage{tcolorbox}
\usepackage[noend]{algpseudocode}
\definecolor{Gray}{gray}{0.92}
\definecolor{LightCyan}{rgb}{0.92,0.968,0.968}
\usepackage{todonotes}
\newcommand{\notcheckmark}{{$\checkmark$}\textsuperscript{\textcolor{black}{\kern-0.65em{\bf--}}}}

\usepackage{CJKutf8}

\usepackage{xcolor}
\definecolor{myred}{rgb}{0.6, 0.15, 0.3}
\usepackage{todonotes}
\makeatletter
\newcommand*\iftodonotes{\if@todonotes@disabled\expandafter\@secondoftwo\else\expandafter\@firstoftwo\fi}
\makeatother

\newcommand\rurl[1]{%
  \href{http://#1}{\nolinkurl{#1}}%
}

\newcommand{\rparagraph}[1]{\vspace{1.6mm}\noindent\textbf{#1.}}
\newcommand{\rparagraphnodot}[1]{\vspace{1.6mm}\noindent\textbf{#1}}
\newcommand{\sparagraph}[1]{\vspace{0.0mm}\noindent\textbf{#1.}}
\newcommand{\sparagraphnodot}[1]{\vspace{0.0mm}\noindent\textbf{#1}}

\newcolumntype{Y}{>{\centering\arraybackslash}X}

\newcommand{\vecmap}{\textsc{\footnotesize VecMap}\xspace}
\newcommand{\contrastivebli}{\textsc{\footnotesize ContrastiveBLI}\xspace}
\newcommand{\sail}{\textsc{\footnotesize SAIL}\xspace}

\newcommand{\chatgpt}{\textsc{\footnotesize ChatGPT}\xspace}
\newcommand{\gpt}{\textsc{\footnotesize GPT}\xspace}

\newcommand{\mapping}{\textsc{\footnotesize Mapping-Based}\xspace}
\newcommand{\zero}{\textsc{\footnotesize Zero-Shot}\xspace}
\newcommand{\llama}{\textsc{\footnotesize LLaMA}\xspace}

\title{Self-Augmented In-Context Learning for Unsupervised Word Translation}

\author{Yaoyiran Li \quad Anna Korhonen \quad Ivan Vuli\'{c}
\\ Language Technology Lab, TAL, University of Cambridge \\
 \texttt{\{yl711,alk23,iv250\}@cam.ac.uk}\\} 

\begin{document}
\maketitle
\begin{abstract}
Recent work has shown that, while large language models (LLMs) demonstrate strong word translation or bilingual lexicon induction (BLI) capabilities in few-shot setups, they still cannot match the performance of `traditional' mapping-based approaches in the unsupervised scenario where no seed translation pairs are available, especially for lower-resource languages. To address this challenge with LLMs, we propose \textbf{s}elf-\textbf{a}ugmented \textbf{i}n-context \textbf{l}earning (\sail) for unsupervised BLI: starting from a zero-shot prompt, \sail iteratively induces a set of high-confidence word translation pairs for in-context learning (ICL) from an LLM, which it then reapplies to the same LLM in the ICL fashion. Our method shows substantial gains over zero-shot prompting of LLMs on two established BLI benchmarks spanning a wide range of language pairs, also outperforming mapping-based baselines across the board. In addition to achieving state-of-the-art unsupervised BLI performance, we also conduct comprehensive analyses on \sail and discuss its limitations.
\end{abstract}

\section{Introduction and Motivation}
\label{s:introduction}
The task of word translation (WT), also known as bilingual lexicon induction (BLI), aims to automatically induce lexica of words with the same or similar meaning in different languages, thus bridging the lexical gap between languages. Even in the era of large language models (LLMs), BLI still has wide applications in machine translation and cross-lingual transfer learning~\citep{sun-etal-2021-cross,zhou-etal-2021-improving,wang-etal-2022-expanding,ghazvininejad2023dictionary,jones2023bilex}. A particular BLI setup, termed \textit{(fully) unsupervised BLI}, is especially compelling because it is not only more technically challenging but is also used as a pivotal component towards unsupervised machine translation \citep{conneau2017word,artetxe2018unsupervised,marchisio-etal-2020-unsupervised,chronopoulou-etal-2021-improving}. 

Until recently, BLI approaches have predominantly relied on learning cross-lingual word embedding (CLWE) mappings: these are known as \mapping approaches and are developed based on static or decontextualised word embeddings (WEs)~\citep{patra-etal-2019-bilingual,pmlr-v89-grave19a,li-etal-2022-improving,10.1007/978-3-031-44696-2_30}. Meanwhile, autoregressive LLMs have become the cornerstone of modern NLP techniques~\citep{NEURIPS2020_1457c0d6,NEURIPS2022_b1efde53,touvron2023llama} with success in many real-world tasks~\citep{KASNECI2023102274,wu2023unleashing,thirunavukarasu2023large,li2024calrec}. Given this trend, recent BLI research has also started to shift towards exploring LLMs. \citet{li-etal-2023-bilingual} first show that prompting LLMs with gold-standard WT pairs as in-context examples (few-shot in-context learning: ICL) outperforms all existing BLI approaches in the supervised and semi-supervised BLI setups (where typically 1K$\sim$5K gold-standard WT pairs are available for training or ICL), while zero-shot prompting still falls behind traditional \mapping approaches in the fully unsupervised BLI setup, especially for lower-resource languages.

In this work, we thus aim at improving unsupervised BLI with LLMs. To this end, we analyze the limitations of zero-shot prompting and propose a novel \textbf{s}elf-\textbf{a}ugmented \textbf{i}n-context \textbf{l}earning (\sail) method for unsupervised BLI with LLMs. The key idea is to first retrieve a set of high-confidence WT pairs by zero-shot prompting LLMs, then iteratively refine the high-confidence dictionary and finally use the gradually refined bilingual lexicon for BLI inference in an ICL fashion (\S\ref{s:methodology}). Our extensive experiments show that \sail establishes new state-of-the-art unsupervised BLI performance on two standard BLI benchmarks. We also conduct thorough analyses on our approach, providing further insights into its inner workings (\S\ref{s:experiments}-\S\ref{s:results}). Our code is publicly available at \url{https://github.com/cambridgeltl/sail-bli}.

\section{Methodology}
\label{s:methodology}
\sparagraph{Unsupervised BLI: Task Preliminaries} We assume a pair of two languages: a source language $L^x$ with its vocabulary $\mathcal{X}$ and a target language $L^y$ with vocabulary $\mathcal{Y}$. In a typical, standard BLI setup the vocabulary of each language contains the most frequent $200,000$ word types in the language~\citep{glavas-etal-2019-properly,li-etal-2022-improving}. Given a source word $w^x\in\mathcal{X}$, the unsupervised BLI task then aims to infer its translation in $L^y$, without any word-level parallel data (i.e., seed translation pairs from a lexicon) available for training or ICL.\footnote{Following prior work, when $w^x$ has multiple ground truth translations in $L^y$, a prediction is considered correct if it is any of the ground truth answers.}

\rparagraph{Zero-Shot Prompting}~\citet{li-etal-2023-bilingual} have proposed to prompt autoregressive LLMs for the BLI task, where the input word $w^x$ is embedded into a predefined text template. We adopt the pool of templates provided by~\citet{li-etal-2023-bilingual} and conduct template search for each LLM on a randomly chosen language pair. As an example, the zero-shot template for \llama-2$_{\text{7B}}$ is as follows:\footnote{The full list of templates used for other LLMs are presented in Appendix~\ref{appendix:templates}.}
\begin{quote}
`\verb|The |$L^x$\verb| word |$w^{x}$\verb| in |$L^y$\verb| is:|',
\end{quote}
where $L^x$, $L^y$, and $w^{x}$ are placeholders for the source language, target language, and the query word in the source language (e.g., $L^x$ = \verb|Hungarian|, $w^{x}$ = \verb|macska|, $L^y$ = \verb|Catalan|).

The deterministic beam search (with beam size of $n$ as a hyper-parameter) is adopted to generate $n$ output text pieces in the final beam, ranked by their sequence scores.\footnote{We use $n=5$ following~\citet{li-etal-2023-bilingual}.} For each of the $n$ outputs, the first word in the generated output following the input sequence is extracted as a candidate answer. After filtering out those candidate answers not in $\mathcal{Y}$, the candidate $L^y$ word with the highest associated sequence score is returned as the final word translation prediction.

\rparagraph{Limitations of Zero-Shot Prompting} The above zero-shot approach for unsupervised BLI, proposed by~\citet{li-etal-2023-bilingual}, comes with several limitations. First, the template does not stipulate the output format and thus parsing the output text may not be as straightforward as expected. Put simply, LLM's prediction may not be the first word in the generated sequence. Second, the LLM may not fully `understand' the input template and sometimes may tend not to generate words in the target language especially for lower-resource languages. For the \textit{supervised} BLI setup, where a dictionary of gold standard translation pairs is assumed and available, few-shot in-context learning can substantially improve final BLI performance~\citep{li-etal-2023-bilingual}, since it not only provides examples of the desired output format but also helps LLMs `understand' the BLI task. However, the availability of such a seed dictionary is not assumed in the \textit{unsupervised} BLI task variant, and the key idea of this work is to derive and iteratively refine a seed dictionary by prompting LLMs.

\rparagraph{\sail: Self-Augmented In-Context Learning for Unsupervised BLI} 
We thus propose to facilitate and improve unsupervised BLI by \textbf{S1)} using zero-shot prompting to retrieve $\mathcal{D}_h$, a set of high-confidence translation pairs, and then \textbf{S2)} leveraging these pairs as `self-augmented' in-context examples for few-shot prompting to further iteratively refine $\mathcal{D}_h$ (across 0 to $N_{it}-1$ iterations, where $N_{it}$ is a hyper-parameter denoting total times of $\mathcal{D}_h$ inference in S1 and S2), and finally \textbf{S3)} conducting few-shot learning with the final, $N_{it}$-th self-created seed lexicon $\mathcal{D}_h$ for BLI inference on the test set.

\rparagraph{Deriving High-Confidence Pairs} For both steps S1 and S2 outlined above, we start with the most frequent $N_{f}$ words in $L^x$ since representations of less frequent words are considered to be much noisier in general~\citep{artetxe-etal-2018-robust}. For each $w^x$, we conduct $L^x\to L^y$ translation: we refer to this predicted word as $\hat{w}^y$.\footnote{We do \textit{not} require $\hat{w}^y$ to be one of the most frequent $N_{f}$ words in $L^y$.} We then propose to conduct \textit{word back-translation}, translating $\hat{w}^y$ from $L^y$ back into $L^x$. The word pair ($w^x$, $\hat{w}^y$) is considered a high-confidence pair only if $w^x$ is also the output word of the back-translation step.\footnote{Earlier \mapping approaches have retrieved high-confidence pairs through ranking cross-lingual word similarity scores (e.g., cosine similarity) to refine CLWE mappings~\citep{artetxe-etal-2018-robust,li-etal-2022-improving}; in a sense, our work renovates and revitalises the idea with LLMs.} We denote the set of all high-confidence pairs from the $L^x$ words as $\mathcal{D}_h^x$. Likewise, we also start from the most frequent $N_{f}$ words in $L^y$ and symmetrically derive $\mathcal{D}_h^y$. Finally, we update the high-confidence dictionary with $\mathcal{D}_h=\mathcal{D}_h^x\cup\mathcal{D}_h^y$.\footnote{Therefore, $|\mathcal{D}_h^x|\leq N_{f}$, $|\mathcal{D}_h^y|\leq N_{f}$, and $|\mathcal{D}_h|\leq2\times N_{f}$.}  


\rparagraph{Few-Shot Prompting with High-Confidence Pairs}
Step S1 of \sail relies on zero-shot prompting, but all the subsequent iterations in S2 and S3 apply few-shot prompting/ICL with the `self-augmented' high-confidence translation pairs $\mathcal{D}_h$. Following~\citet{li-etal-2023-bilingual}, we adopt 5-shot prompting, and again conduct template search on the BLI task with a single, randomly selected language pair.\footnote{The decoding and output parsing strategy is the same as in zero-shot prompting.} The in-context examples, $(w_{i}^{x},w_{i}^{y})\in\mathcal{D}_h,1\leq i\leq 5$, are retrieved where the $w_{i}^{x}$ words are the nearest neighbours of the input word $w^x$ in $L^x$'s static word embedding space. The few-shot template for \llama-2$_{\text{7B}}$ is then as follows:
\begin{quote}
`\verb|The |$L^x$\verb| word |$w_{1}^{x}$\verb| in |$L^y$\verb| is |$w_{1}^{y}$\verb|. The |\\$L^x$\verb| word |$w_{2}^{x}$\verb| in |$L^y$\verb| is |$w_{2}^{y}$\verb|. ... The |$L^x$\\\verb|word |$w^{x}$\verb| in |$L^y$\verb| is|'.
\end{quote}

\section{Experimental Setup}
\label{s:experiments}
\sparagraph{BLI Data and LLMs}
We adopt two standard BLI benchmarks: \textbf{1)} $5$ languages from XLING~\citep{glavas-etal-2019-properly} including German (\textsc{de}), English (\textsc{en}), French (\textsc{fr}), Italian (\textsc{it}), and Russian (\textsc{ru}), their combinations resulting in $20$ BLI directions; \textbf{2)} $3$ lower-resource languages including Bulgarian (\textsc{bg}), Catalan (\textsc{ca}), and Hungarian (\textsc{hu}) from PanLex-BLI~\citep{vulic-etal-2019-really}, which result in $6$ BLI directions.\footnote{The two datasets are also used in many recent BLI works~\citep{sachidananda2021filtered,aboagye2022normalization,li-etal-2022-improving,li-etal-2022-improving-bilingual,vulic-etal-2020-probing,vulic-etal-2023-probing,li-etal-2023-bilingual}.} For both benchmarks, a test set of $2$K WT pairs is provided for each BLI direction. We experiment with four open-source LLMs: \llama$_{\text{7B}}$, \llama-2$_{\text{7B}}$, \llama$_{\text{13B}}$, and \llama-2$_{\text{13B}}$~\citep{touvron2023llama,touvron2023llama2}.~\citet{li-etal-2023-bilingual} found that $4$ other families of LLMs, including mT5, mT0, mGPT and XGLM, underperform \llama; we thus skip these LLMs in our work. 

\rparagraph{Implementation Details and BLI Evaluation} As mentioned in \S\ref{s:methodology}, our hyper-parameter and template search are conducted on a single, randomly selected language pair, which is \textsc{de}-\textsc{fr}, following~\citet{li-etal-2023-bilingual}. Batch size is set to $1$. We adopt $N_{it}=1$, $N_{f}=5,000$ in our main experiments (\S\ref{r:main}) and then investigate their influence on BLI performance and the effectiveness of our proposed word back-translation in our further analyses (\S\ref{r:fa}). Half-precision floating-point format (\verb|torch.float16|) is adopted for all our \sail and zero-shot experiments. Since our method does \textit{not} imply any randomness, 
all results are from single runs. For evaluation, we adopt the standard \textit{top-1 accuracy} as prior work.

\rparagraph{Baselines} We adopt two established \mapping baselines. \textbf{1)} \vecmap is a representative unsupervised BLI approach and features a self-learning mechanism that refines linear maps for deriving CLWEs~\citep{artetxe-etal-2018-robust}. \textbf{2)} \contrastivebli learns CLWEs with a two-stage contrastive learning framework and is the strongest \mapping approach for supervised and semi-supervised BLI tasks on our two benchmarks~\citep{li-etal-2022-improving}; however, it does not support unsupervised setup. We extend \contrastivebli to unsupervised BLI by initialising the initial map with the unsupervised \vecmap method. The \contrastivebli C1 variant based on static WEs and its stronger C2 variant combining static and decontextualised WEs are both used as our baselines. We adopt Cross-domain Similarity Local Scaling (CSLS) retrieval~\cite{conneau2017word} for all \mapping approaches as recommended in the baselines. In addition, we report \textbf{3)} \zero prompting with each of our LLMs as baselines following the previous findings of \citet{li-etal-2023-bilingual}.

\section{Results and Discussion}
\label{s:results}
\subsection{Main Results}
\label{r:main}

\sparagraphnodot{Results on the Two BLI Benchmarks} are summarised in Tables~\ref{table:mainres} and~\ref{table:mainresp} respectively, with full BLI scores per each individual language pair in Tables~\ref{table:mainresappendix} and~\ref{table:mainrespappendix} in Appendix~\ref{appendix:res}. As the main findings, \textbf{1)} our \sail shows consistent gains against \zero prompting for each of the $4$ LLMs, showing the effectiveness of the proposed approach; \textbf{2)} while \zero prompting still lags behind \mapping approaches on PanLex-BLI's lower-resource languages, applying \sail outperforms \mapping baselines across the board. The only exception is that \contrastivebli (C2) still has a slight edge over \sail with the weakest LLM overall, \llama$_{\text{7B}}$. \textbf{3)} Among the $4$ LLMs, \llama-2$_{\text{13B}}$ presents the strongest BLI capability.

\begin{table}[!t]
\def\arraystretch{0.995}
\begin{center}
\resizebox{0.48\textwidth}{!}{%
\begin{tabular}{lllllll}
\toprule 
\rowcolor{Gray}
\multicolumn{1}{c}{\bf [Unsupervised BLI]} &\multicolumn{1}{c}{\bf \textsc{de}} &\multicolumn{1}{c}{\bf \textsc{en}}
&\multicolumn{1}{c}{\bf \textsc{fr}} &\multicolumn{1}{c}{\bf \textsc{it}} &\multicolumn{1}{c}{\bf \textsc{ru}} &\multicolumn{1}{c}{\bf \textsc{avg.}} \\ 
\midrule
& \multicolumn{6}{c}{\bf \mapping}\\
\cmidrule(lr){2-7}
\multicolumn{1}{c}{\vecmap} &\multicolumn{1}{c}{44.14}&\multicolumn{1}{c}{51.7}&\multicolumn{1}{c}{51.51}&\multicolumn{1}{c}{51.03}&\multicolumn{1}{c}{34.36}&\multicolumn{1}{c}{46.55}\\
\multicolumn{1}{c}{\contrastivebli (C1)} &\multicolumn{1}{c}{44.72}&\multicolumn{1}{c}{52.12}&\multicolumn{1}{c}{52.29}&\multicolumn{1}{c}{51.77}&\multicolumn{1}{c}{35.5}&\multicolumn{1}{c}{47.28}\\
\multicolumn{1}{c}{\contrastivebli (C2)} &\multicolumn{1}{c}{46.02}&\multicolumn{1}{c}{53.32}&\multicolumn{1}{c}{53.26}&\multicolumn{1}{c}{52.99}&\multicolumn{1}{c}{37.26}&\multicolumn{1}{c}{48.57}\\
\midrule
& \multicolumn{6}{c}{\bf \zero}\\
\cmidrule(lr){2-7}
\multicolumn{1}{c}{\llama$_{\text{7B}}$} &\multicolumn{1}{c}{41.94}&\multicolumn{1}{c}{50.16}&\multicolumn{1}{c}{48.25}&\multicolumn{1}{c}{46.91}&\multicolumn{1}{c}{40.04}&\multicolumn{1}{c}{45.46}\\
\multicolumn{1}{c}{\llama-2$_{\text{7B}}$} &\multicolumn{1}{c}{43.91}&\multicolumn{1}{c}{52.7}&\multicolumn{1}{c}{50.68}&\multicolumn{1}{c}{48.23}&\multicolumn{1}{c}{42.8}&\multicolumn{1}{c}{47.66}\\
\multicolumn{1}{c}{\llama$_{\text{13B}}$} &\multicolumn{1}{c}{45.39}&\multicolumn{1}{c}{53.35}&\multicolumn{1}{c}{52.39}&\multicolumn{1}{c}{50.58}&\multicolumn{1}{c}{41.74}&\multicolumn{1}{c}{48.69}\\
\multicolumn{1}{c}{\llama-2$_{\text{13B}}$} &\multicolumn{1}{c}{47.12}&\multicolumn{1}{c}{55.02}&\multicolumn{1}{c}{51.31}&\multicolumn{1}{c}{52.02}&\multicolumn{1}{c}{43.09}&\multicolumn{1}{c}{49.71}\\
\midrule
& \multicolumn{6}{c}{\bf \sail (Ours)}\\
\cmidrule(lr){2-7}
\multicolumn{1}{c}{\llama$_{\text{7B}}$} &\multicolumn{1}{c}{51.39}&\multicolumn{1}{c}{61.92}&\multicolumn{1}{c}{58.92}&\multicolumn{1}{c}{56.94}&\multicolumn{1}{c}{50.7}&\multicolumn{1}{c}{55.97}\\
\multicolumn{1}{c}{\llama-2$_{\text{7B}}$} &\multicolumn{1}{c}{53.81}&\multicolumn{1}{c}{64.12}&\multicolumn{1}{c}{61.09}&\multicolumn{1}{c}{59.96}&\multicolumn{1}{c}{53.77}&\multicolumn{1}{c}{58.55}\\
\multicolumn{1}{c}{\llama$_{\text{13B}}$} &\multicolumn{1}{c}{55.35}&\multicolumn{1}{c}{64.84}&\multicolumn{1}{c}{62.49}&\multicolumn{1}{c}{61.27}&\multicolumn{1}{c}{54.5}&\multicolumn{1}{c}{59.69}\\
\multicolumn{1}{c}{\llama-2$_{\text{13B}}$} &\multicolumn{1}{c}{\bf 57.69}&\multicolumn{1}{c}{\bf 67.0}&\multicolumn{1}{c}{\bf 64.11}&\multicolumn{1}{c}{\bf 63.18}&\multicolumn{1}{c}{\bf 57.04}&\multicolumn{1}{c}{\bf 61.8}\\

\bottomrule
\end{tabular}
}
\caption{Main results on the $20$ XLING BLI directions. For each language, the average accuracy scores over $8$ BLI directions (i.e., going from and going to other $4$ languages) is reported. See also Appendix~\ref{appendix:res}.}
\label{table:mainres}
\end{center}
\end{table}

\begin{table}[!t]
\def\arraystretch{0.999}
\begin{center}
\resizebox{0.48\textwidth}{!}{%
\begin{tabular}{lllll}
\toprule 
\rowcolor{Gray}
\multicolumn{1}{c}{\bf [Unsupervised BLI]} &\multicolumn{1}{c}{\bf \textsc{bg}} &\multicolumn{1}{c}{\bf \textsc{ca}}
&\multicolumn{1}{c}{\bf \textsc{hu}} &\multicolumn{1}{c}{\bf \textsc{avg.}} \\ 
\midrule
& \multicolumn{4}{c}{\bf \mapping}\\
\cmidrule(lr){2-5}
\multicolumn{1}{c}{\vecmap} &\multicolumn{1}{c}{37.22}&\multicolumn{1}{c}{36.27}&\multicolumn{1}{c}{36.89}&\multicolumn{1}{c}{36.8}\\
\multicolumn{1}{c}{\contrastivebli (C1)} &\multicolumn{1}{c}{36.7}&\multicolumn{1}{c}{35.86}&\multicolumn{1}{c}{37.82}&\multicolumn{1}{c}{36.79}\\
\multicolumn{1}{c}{\contrastivebli (C2)} &\multicolumn{1}{c}{38.87}&\multicolumn{1}{c}{38.48}&\multicolumn{1}{c}{40.54}&\multicolumn{1}{c}{39.3}\\
\midrule
& \multicolumn{4}{c}{\bf \zero}\\
\cmidrule(lr){2-5}
\multicolumn{1}{c}{\llama$_{\text{7B}}$} &\multicolumn{1}{c}{27.9}&\multicolumn{1}{c}{28.87}&\multicolumn{1}{c}{27.18}&\multicolumn{1}{c}{27.98}\\
\multicolumn{1}{c}{\llama-2$_{\text{7B}}$} &\multicolumn{1}{c}{28.2}&\multicolumn{1}{c}{27.21}&\multicolumn{1}{c}{26.92}&\multicolumn{1}{c}{27.45}\\
\multicolumn{1}{c}{\llama$_{\text{13B}}$} &\multicolumn{1}{c}{27.49}&\multicolumn{1}{c}{30.61}&\multicolumn{1}{c}{28.2}&\multicolumn{1}{c}{28.77}\\
\multicolumn{1}{c}{\llama-2$_{\text{13B}}$} &\multicolumn{1}{c}{29.08}&\multicolumn{1}{c}{32.38}&\multicolumn{1}{c}{30.53}&\multicolumn{1}{c}{30.66}\\
\midrule
& \multicolumn{4}{c}{\bf \sail (Ours)}\\
\cmidrule(lr){2-5}
\multicolumn{1}{c}{\llama$_{\text{7B}}$} &\multicolumn{1}{c}{37.02}&\multicolumn{1}{c}{37.63}&\multicolumn{1}{c}{36.29}&\multicolumn{1}{c}{36.98}\\
\multicolumn{1}{c}{\llama-2$_{\text{7B}}$} &\multicolumn{1}{c}{40.06}&\multicolumn{1}{c}{40.51}&\multicolumn{1}{c}{40.22}&\multicolumn{1}{c}{40.27}\\
\multicolumn{1}{c}{\llama$_{\text{13B}}$} &\multicolumn{1}{c}{41.71}&\multicolumn{1}{c}{42.76}&\multicolumn{1}{c}{42.07}&\multicolumn{1}{c}{42.18}\\
\multicolumn{1}{c}{\llama-2$_{\text{13B}}$} &\multicolumn{1}{c}{\bf 45.4}&\multicolumn{1}{c}{\bf 46.26}&\multicolumn{1}{c}{\bf 44.88}&\multicolumn{1}{c}{\bf 45.51}\\

\bottomrule
\end{tabular}
}
\caption{Main results on $6$ PanLex-BLI BLI directions. For each language, the average accuracy scores over $4$ BLI directions (i.e., going from and going to other $2$ languages) is reported. See also Appendix~\ref{appendix:res}.}
\label{table:mainresp}
\end{center}
\end{table}

\rparagraph{Variance and Statistical Significance} The whole \sail method does \textit{not} imply any variance due to randomness: it does not rely on any actual LLM  fine-tuning; we adopt deterministic beam search; the deterministic nearest neighbour retrieval is used for deriving in-context examples. Here, we report the statistical significance with $\chi^{2}$ tests. When comparing \sail and \zero prompting (both with \llama-2$_{\text{13B}}$), the $p$-value is $1.1e$-$251$ on 20 XLING BLI directions and $2.7e$-$109$ on 6 PanLex-BLI BLI directions. We then compare \sail (with \llama-2$_{\text{13B}}$) against \contrastivebli (C2) which is our strongest \mapping baseline: the $p$-values are $3.1e$-$300$ and $7.8e$-$20$ respectively. These show that our findings are strongly statistically significant.\footnote{Usually $p<0.05$ or $p<0.001$ is considered to indicate statistical significance.}

\subsection{Further Analyses}
\label{r:fa}

\sparagraph{Inspection of High-Confidence Dictionaries} To provide additional insight into our \sail approach, we present statistics on the size of high-confidence dictionaries derived in our main experiments ($N_{it}=1$, $N_{f}=5,000$, and with word back-translation) over $20$ XLING BLI directions and $6$ PanLex-BLI BLI directions respectively for each of our four LLMs in Table~\ref{table:dictionarysizes}. The values indicate that $|\mathcal{D}_h|$ of higher-resource languages (XLING) is typically greater than that of lower-resource languages (PanLex-BLI). In addition to the dictionary size, it is also worth investigating the quality of high-confidence dictionaries. However, to directly evaluate the quality of the ‘silver standard’ generated dictionaries is difficult since we do not have ground truth dictionaries for comparison. As a preliminary investigation, we randomly sample $50$ translation pairs from the \textsc{en}-\textsc{de} \llama-2$_{\text{13B}}$-augmented dictionary and compare them with answers derived from Google Translate\footnote{\url{https://translate.google.com/}} (\textsc{en}$\to$\textsc{de}). We found that $40$ out of the $50$ pairs in our augmented dictionary are the same as the results from Google Translate. Although these results from Google Translate are also not ‘gold standard’ ground truth, it does point in the direction of reliability of extracted WT pairs.

\begin{table}[!t]
\def\arraystretch{1.0}
\begin{center}
\resizebox{0.48\textwidth}{!}{%
\begin{tabular}{lllll}
\toprule 
\rowcolor{Gray}
\multicolumn{1}{c}{\bf LLM (\sail)} &\multicolumn{2}{c}{\bf $|\mathcal{D}_h|$: XLING} &\multicolumn{2}{c}{\bf $|\mathcal{D}_h|$: PanLex-BLI}\\

\cmidrule(lr){2-3} \cmidrule(lr){4-5}

\multicolumn{1}{c}{} &\multicolumn{1}{c}{\bf \textsc{mean}} &\multicolumn{1}{c}{\bf\textsc{min}$\sim$\textsc{max}} &\multicolumn{1}{c}{\bf\textsc{mean}} &\multicolumn{1}{c}{\bf\textsc{min}$\sim$\textsc{max}}\\

\multicolumn{1}{c}{\llama$_{\text{7B}}$} &\multicolumn{1}{c}{2471} &\multicolumn{1}{c}{1731$\sim$3180} &\multicolumn{1}{c}{1735} &\multicolumn{1}{c}{1363$\sim$2095}\\

\multicolumn{1}{c}{\llama-2$_{\text{7B}}$} &\multicolumn{1}{c}{3019} &\multicolumn{1}{c}{2086$\sim$3824} &\multicolumn{1}{c}{1873} &\multicolumn{1}{c}{1690$\sim$2183}\\

\multicolumn{1}{c}{\llama$_{\text{13B}}$} &\multicolumn{1}{c}{2850} &\multicolumn{1}{c}{	2064$\sim$3579} &\multicolumn{1}{c}{2005} &\multicolumn{1}{c}{1548$\sim$2351}\\

\multicolumn{1}{c}{\llama-2$_{\text{13B}}$} &\multicolumn{1}{c}{2612} &\multicolumn{1}{c}{1577$\sim$3362} &\multicolumn{1}{c}{1737} &\multicolumn{1}{c}{1184$\sim$2049}\\

\bottomrule
\end{tabular}
}
\caption{Statistics on $|\mathcal{D}_h|$ for each LLM over $20$ XLING BLI directions and $6$ PanLex-BLI BLI directions respectively.}
\label{table:dictionarysizes}
\end{center}
\end{table}

\begin{figure}[!t]
    \centering
    \includegraphics[width=0.98\linewidth]{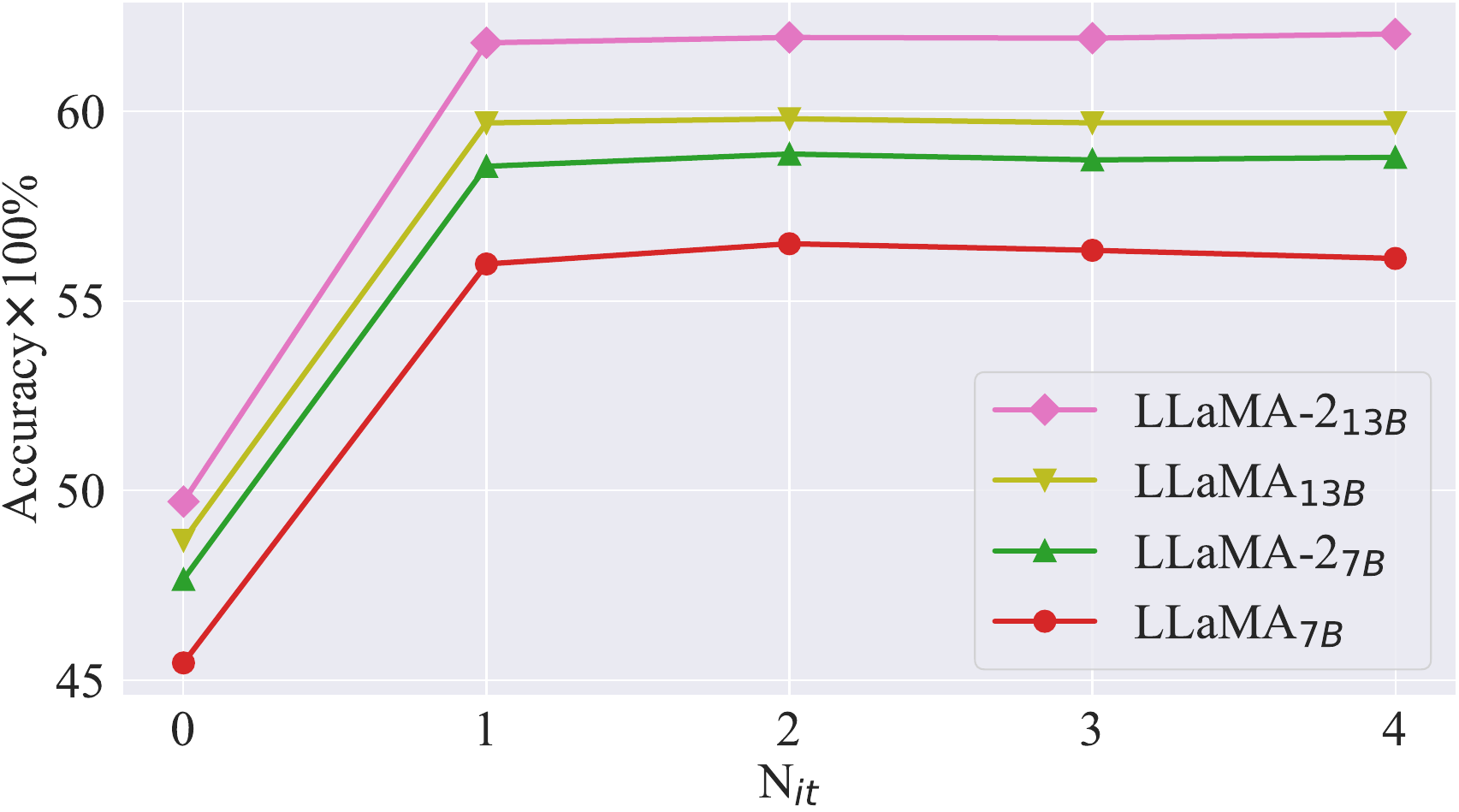}
    \caption{Top-1 accuracy ($\times 100\%$) averaged over $20$ XLING BLI directions with respect to $N_{it}$. $N_{it}=0$ yields the \zero baseline.}
    \label{fig:nit}
\end{figure}

\rparagraph{Impact of $N_{it}$} Figure~\ref{fig:nit} shows the influence of the number of iterations $N_{it}$ on the average BLI scores on XLING. When $N_{it}=1$, where only step S1 is executed (see \S\ref{s:methodology}), \sail already approaches (almost) its optimal performance. Further refining the $\mathcal{D}_h$ for more iterations (step S2) only leads to small fluctuations in BLI performance, which we deem not worth the increased computational cost. Figure~\ref{fig:nitp} (Appendix~\ref{appendix:nit}) with results on PanLex-BLI shows a similar trend.

\begin{figure}[!t]
    \centering
    \includegraphics[width=0.98\linewidth]{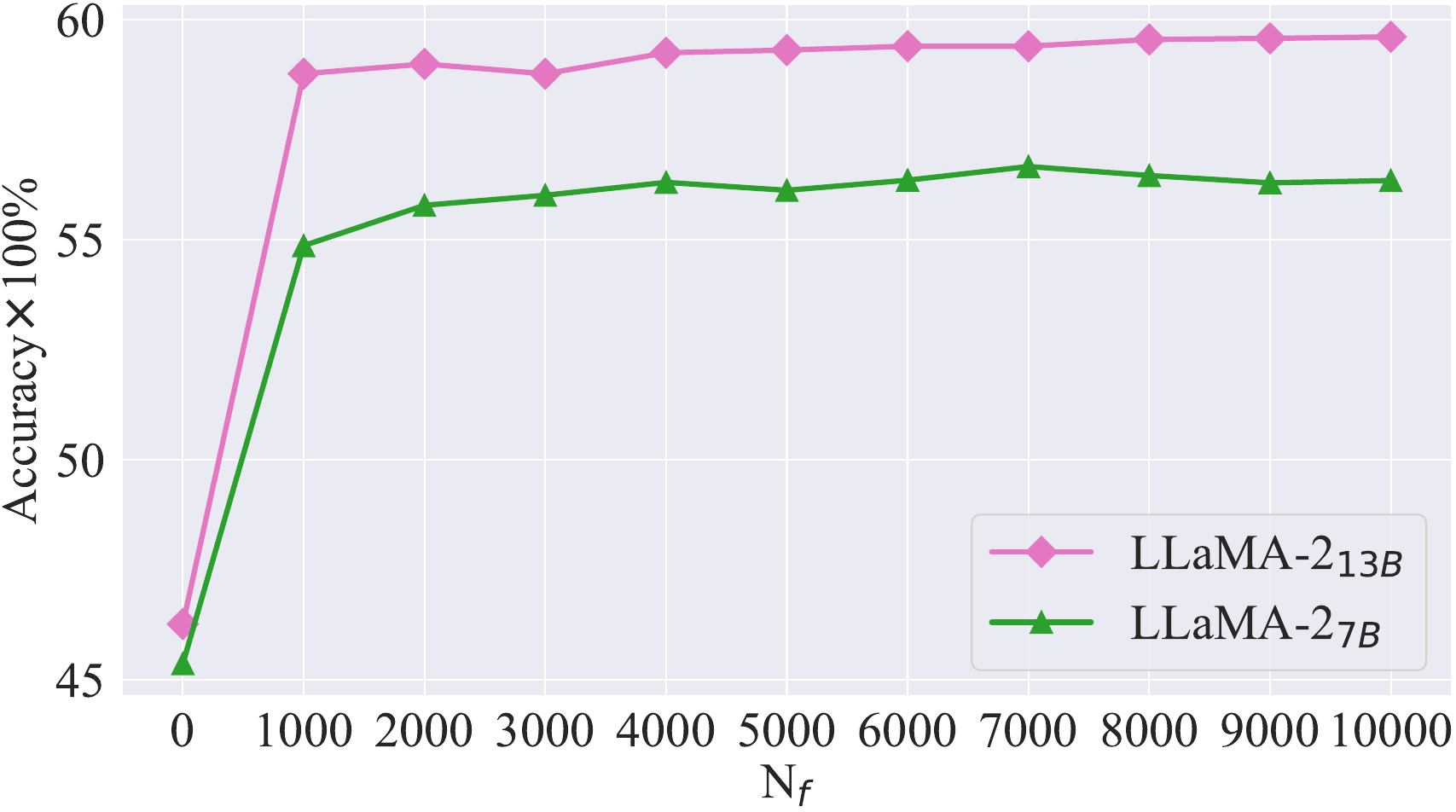}
    \caption{Top-1 accuracy on a subset of XLING with respect to $N_{f}$. $N_{f}=0$ yields the \zero baseline.}
    \label{fig:nf}
\end{figure}

\rparagraph{Impact of $N_{f}$} We then study the impact of the frequency threshold $N_{f}$ on the average BLI performance with a subset of XLING spanning \textsc{de}-\textsc{fr}, \textsc{en}-\textsc{ru} and \textsc{ru}-\textsc{fr}, each in both directions. The results in Figure~\ref{fig:nf} reveal that even with $N_{f}=1,000$, the BLI performance is boosted substantially when compared against the \zero baseline (i.e., when $N_{f}=0$). When we further increase $N_{f}$, the accuracy score still increases slowly, and the gain seems negligible with $N_{f}\geq5000$: i.e., increasing $N_{f}$ again may not be worth the extra computation. 

\rparagraph{Impact of Word Back-Translation} The back-translation step aims to improve the quality of $\mathcal{D}_h$. Here, we experiment with the ablated version of \sail without back-translation on the same XLING subset (\textsc{de}-\textsc{fr}, \textsc{en}-\textsc{ru} and \textsc{ru}-\textsc{fr}) as before. The results in Table~\ref{table:ablation} clearly demonstrate the effectiveness of proposed word back-translation: the $p$-values ($\chi^{2}$ tests) are $8.8e$-$7$ and $1.0e$-$10$ respectively for \llama-2$_{\text{7B}}$ and \llama-2$_{\text{13B}}$ when comparing \sail variants with and without the back-translation mechanism. 

\begin{table}[!t]
\def\arraystretch{1.0}
\begin{center}
\resizebox{0.48\textwidth}{!}{%
\begin{tabular}{llll}
\toprule 
\rowcolor{Gray}
\multicolumn{1}{c}{\bf LLM} &\multicolumn{1}{c}{\bf\zero} &\multicolumn{1}{c}{\bf\sail(w/o back-translation)} &\multicolumn{1}{c}{\bf\sail}\\

\midrule

\multicolumn{1}{c}{\llama-2$_{\text{7B}}$} &\multicolumn{1}{c}{45.36}&\multicolumn{1}{c}{52.9}&\multicolumn{1}{c}{\bf 56.12}\\

\multicolumn{1}{c}{\llama-2$_{\text{13B}}$} &\multicolumn{1}{c}{46.26}&\multicolumn{1}{c}{55.1}&\multicolumn{1}{c}{\bf 59.31}\\

\bottomrule
\end{tabular}
}
\caption{BLI results on XLING, demonstrating the usefulness of back-translation when constructing $\mathcal{D}_h$. Top-1 accuracy ($\times 100\%$) scores.}
\label{table:ablation}
\end{center}
\end{table}

\rparagraphnodot{\chatgpt for BLI?} We additionally report \gpt-3.5~\citep{chatgpt} and \gpt-4~\citep{achiam2023gpt} results on \textsc{de}-\textsc{fr}, \textsc{en}-\textsc{ru} and \textsc{ru}-\textsc{fr} with \zero prompting (see Appendix~\ref{appendix:chatgpt} for experimental details). Note that the procedure of instruction-tuning of LLMs usually covers large-scale parallel data for machine translation. Therefore, leveraging \chatgpt models, even with \zero prompting, is \textit{not} in line with the motivation of \textit{unsupervised} BLI and leads to unfair comparisons with the results of our main experiments and baselines.\footnote{The four \llama models used in our main experiments are pretrained LLMs without instruction-tuning (see Appendix~\ref{appendix:reproduce}); our \mapping baselines adopt static WEs derived from monolingual corpora of respective languages and our \contrastivebli (C2) baseline additionally leverages pretrained mBERT~\citep{devlin-etal-2019-bert}.} Here, we report \chatgpt results as an upper bound for \zero prompting. Our results in Table~\ref{table:chatgpt} show that \textbf{1)} as expected, the instruction-tuned \chatgpt models outperform pretrained \llama-2$_{\text{13B}}$ by a large margin in the \zero setup, but \textbf{2)} our \sail method with the same pretrained \llama-2$_{\text{13B}}$ outperforms both \gpt-3.5 and the state-of-the-art \gpt-4\footnote{We adopt the strong `\texttt{gpt-4-turbo-2024-04-09}' model which ranked $1^{\text{st}}$ on the \href{https://chat.lmsys.org/?leaderboard}{LMSYS Chatbot Arena Leaderboard} at the time of experimentation (May 12, 2024).} in terms of the average performance, even for the selected higher-resource languages, again demonstrating the effectiveness of the proposed \sail approach.

\begin{table}[!t]
\def\arraystretch{0.87}
\begin{center}
\resizebox{0.47\textwidth}{!}{%
\begin{tabular}{lllll}
\toprule 
\rowcolor{Gray}
\multicolumn{1}{c}{\bf BLI Direction} &\multicolumn{1}{c}{\bf\llama-2$_{\text{13B}}$} &\multicolumn{1}{c}{\bf \gpt-3.5} &\multicolumn{1}{c}{\bf \gpt-4} &\multicolumn{1}{c}{\bf\llama-2$_{\text{13B}}$}\\

\cmidrule(lr){2-4} \cmidrule(lr){5-5}
\multicolumn{1}{c}{} &\multicolumn{3}{c}{\bf\zero} &\multicolumn{1}{c}{\bf\sail}\\

\multicolumn{1}{c}{\textsc{de}$\to$\textsc{fr}}&\multicolumn{1}{c}{46.64} &\multicolumn{1}{c}{59.52} &\multicolumn{1}{c}{\bf62.6} &\multicolumn{1}{c}{61.5}\\

\multicolumn{1}{c}{\textsc{fr}$\to$\textsc{de}} &\multicolumn{1}{c}{50.8} &\multicolumn{1}{c}{58.41} &\multicolumn{1}{c}{\bf60.63} &\multicolumn{1}{c}{56.29}\\

\multicolumn{1}{c}{\textsc{en}$\to$\textsc{ru}} &\multicolumn{1}{c}{47.6} &\multicolumn{1}{c}{55.85} &\multicolumn{1}{c}{55.9} &\multicolumn{1}{c}{\bf63.75}\\

\multicolumn{1}{c}{\textsc{ru}$\to$\textsc{en}} &\multicolumn{1}{c}{51.44} &\multicolumn{1}{c}{59.93} &\multicolumn{1}{c}{\bf60.35} &\multicolumn{1}{c}{59.93}\\

\multicolumn{1}{c}{\textsc{ru}$\to$\textsc{fr}} &\multicolumn{1}{c}{41.17} &\multicolumn{1}{c}{59.77} &\multicolumn{1}{c}{\bf61.39} &\multicolumn{1}{c}{60.29}\\

\multicolumn{1}{c}{\textsc{fr}$\to$\textsc{ru}} &\multicolumn{1}{c}{39.94} &\multicolumn{1}{c}{46.82} &\multicolumn{1}{c}{49.35} &\multicolumn{1}{c}{\bf54.11}\\

\multicolumn{1}{c}{Avg.} &\multicolumn{1}{c}{46.26} &\multicolumn{1}{c}{56.72} &\multicolumn{1}{c}{58.37} &\multicolumn{1}{c}{\bf59.31}\\

\bottomrule
\end{tabular}
}
\vspace{-0.5mm}
\caption{Comparisons with \gpt models.}
\label{table:chatgpt}
\end{center}
\vspace{-4mm}
\end{table}

\section{Conclusion}
\label{s:Conclution}
We proposed Self-Augmented In-Context Learning (\sail) to improve unsupervised BLI with LLMs. The key idea is to iteratively retrieve a set of high-confidence word translation pairs by prompting LLMs and then leverage the retrieved pairs as in-context examples for unsupervised BLI. Our experiments on two standard BLI benchmarks showed that the proposed \sail method substantially outperforms established \mapping and \zero BLI baselines. We also conducted a series of in-depth analyses on the high-confidence dictionary, key hyper-parameters, and the back-translation mechanism, and we additionally show that our \sail approach with \llama-2$_{\text{13B}}$ can even outperform \zero prompting with the state-of-the-art \gpt-4 model.


\section*{Limitations}
\label{s:limitations}
The main limitation of this work, inherited from prior work as well~\cite{li-etal-2023-bilingual} is that the scope of our languages is constrained to the languages supported (or `seen') by the underlying LLMs. For example, \llama-2 is reported to support only around $27$ natural languages~\citep{touvron2023llama2}. This limitation could be mitigated if more advanced LLMs that support more languages are available in the future. It might also be feasible to adapt existing LLMs to more languages by fine-tuning on their monolingual corpora potentially combined with modern cross-lingual transfer learning techniques, whereas such adaptations of LLMs to unseen languages extend way beyond this work focused on the BLI task.

In addition, compared to the \zero baseline, our \sail framework organically requires more computational time and budget, as reported in Table~\ref{table:modelswithsizes} of Appendix~\ref{appendix:reproduce}. 

Moreover, the \sail framework is proposed and evaluated for the unsupervised BLI task. This work does not discuss if and how adapted variants of \sail could also be applied to other NLP tasks beyond BLI. Further, the \sail method should be equally applicable in weakly supervised BLI setups~\cite{vulic-etal-2019-really} where a tiny set of available seed word translations (e.g., 50-500 word pairs) can be assumed to seed the iterative procedure. We leave this to future work.   

\section*{Acknowledgements}
\label{s:acknowledgements}
We thank the anonymous reviewers for their valuable feedback. Yaoyiran Li is supported by Grace $\&$ Thomas C. H. Chan Cambridge International Scholarship. Anna Korhonen is supported by the UK Research and Innovation (UKRI) Frontier Research Grant EP/Y031350/1 (the UK government’s funding guarantee for ERC Advanced Grants). Ivan Vuli\'{c} is supported by a personal Royal Society University Research Fellowship \textit{Inclusive and Sustainable Language Technology for a Truly Multilingual World'} (no 221137).

\bibliography{anthology,custom}
\clearpage
\appendix
\section{Languages}
\label{appendix:languages}

\begin{table}[ht!]
\begin{center}
\resizebox{0.67\linewidth}{!}{%
\begin{tabular}{ccc}
\toprule 
\rowcolor{Gray}
\multicolumn{1}{c}{\bf Family}  &\multicolumn{1}{c}{\bf Language}   &\multicolumn{1}{c}{\bf Code}\\
\midrule
\multirow{2}{*}{Germanic}&\multicolumn{1}{c}{English} &\multicolumn{1}{c}{\textsc{en}} \\
&\multicolumn{1}{c}{German}&\multicolumn{1}{c}{\textsc{de}} \\
\midrule
\multirow{3}{*}{Romance}&\multicolumn{1}{c}{Catalan}&\multicolumn{1}{c}{\textsc{ca}} \\&\multicolumn{1}{c}{French}&\multicolumn{1}{c}{\textsc{fr}}\\
&\multicolumn{1}{c}{Italian}&\multicolumn{1}{c}{\textsc{it}} \\
\midrule
\multirow{2}{*}{Slavic}&\multicolumn{1}{c}{Bulgarian}&\multicolumn{1}{c}{\textsc{bg}} \\&\multicolumn{1}{c}{Russian}  &\multicolumn{1}{c}{\textsc{ru}} \\
\midrule
\multirow{1}{*}{Uralic}&\multicolumn{1}{c}{Hungarian }&\multicolumn{1}{c}{\textsc{hu}} \\
\bottomrule
\end{tabular}
}

\caption{Languages used in our experiments with their ISO 639-1 codes.}
\label{table:languages}
\end{center}
\end{table}

\section{Impact of \texorpdfstring{$N_{it}$}{} with PanLex-BLI}
\label{appendix:nit}
\begin{figure}[ht]
    \centering
    \includegraphics[width=0.9\linewidth]{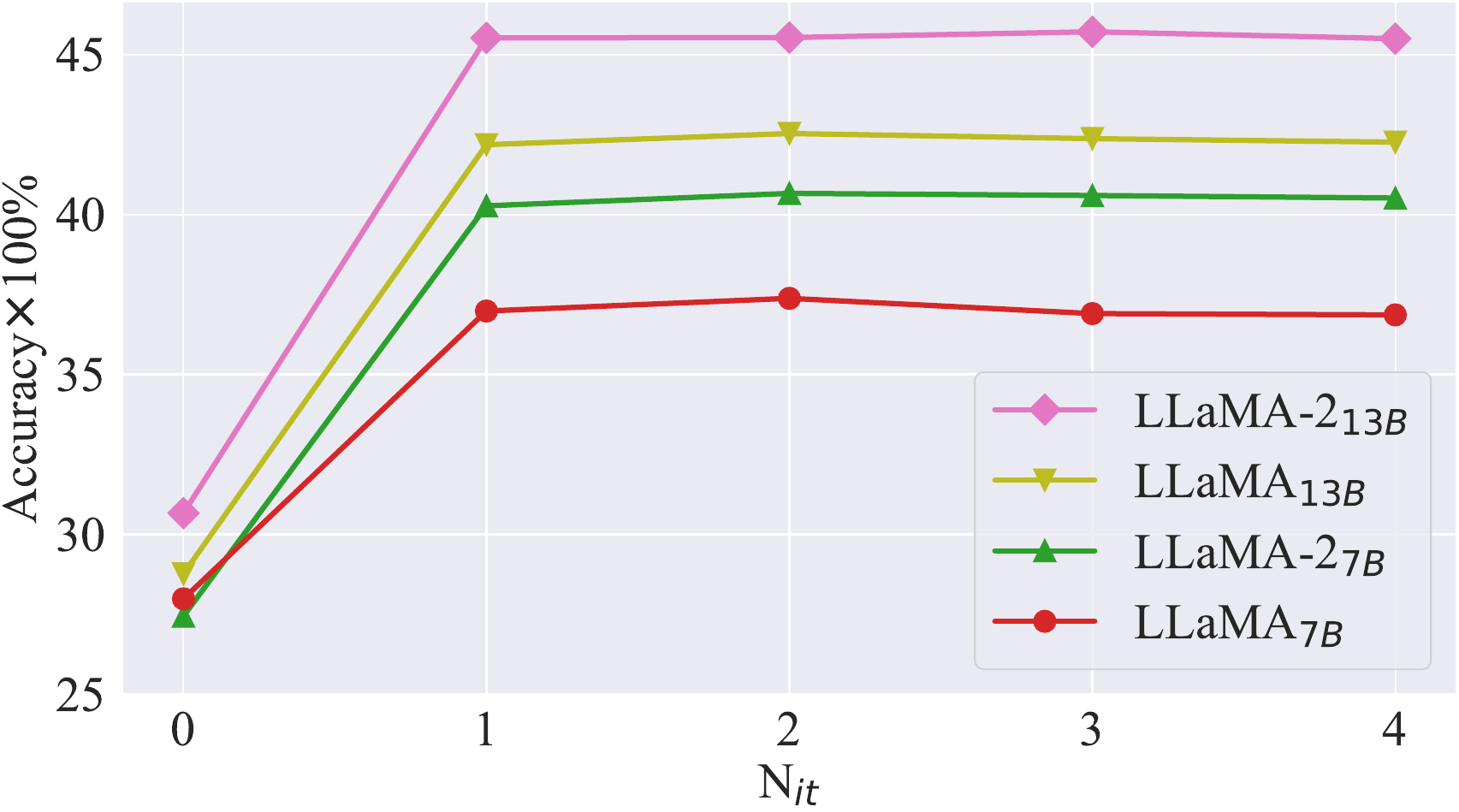}
    \vspace{-1mm}
    \caption{Top-1 accuracy ($\times 100\%$) averaged over $6$ PanLex-BLI BLI directions with respect to $N_{it}$. $N_{it}=0$ yields the \zero baseline.}
    \label{fig:nitp}
    \vspace{-2mm}
\end{figure}

\section{Templates}
\label{appendix:templates}
\citet{li-etal-2023-bilingual} provide the suggested (carefully searched) templates for \llama$_{\text{7B}}$ and \llama$_{\text{13B}}$, which we directly adopt in our work. For \llama-2$_{\text{7B}}$ and \llama-2$_{\text{13B}}$, we conduct template search following \citet{li-etal-2023-bilingual} on a single language pair \textsc{de}-\textsc{fr} in both directions. For \chatgpt models used in \S\ref{r:fa}, details about their templates are provided in Appendix~\ref{appendix:chatgpt}.

\rparagraph{Zero-Shot Template} \llama$_{\text{7B}}$, \llama-2$_{\text{7B}}$ and \llama-2$_{\text{13B}}$ share the same zero-shot template as introduced in \S\ref{s:methodology}. \llama$_{\text{13B}}$'s zero-shot template is as follows:
\begin{quote}
`\verb|Translate from |$L^x$\verb| to |$L^y$\verb|: |$w^{x}$\verb|=>|'.
\end{quote}
\sparagraph{Few-Shot Template}. We have introduced the few-shot template of \llama-2$_{\text{7B}}$ in \S\ref{s:methodology}. The remaining three LLMs happen to share the same few-shot template, given as follows:  
\begin{quote}
`\verb|The |$L^x$\verb| word '|$w_{1}^{x}$\verb|' in |$L^y$\verb| is |$w_{1}^{y}$\verb|. The |\\$L^x$\verb| word '|$w_{2}^{x}$\verb|' in |$L^y$\verb| is |$w_{2}^{y}$\verb|. ... The |$L^x$\\\verb|word '|$w^{x}$\verb|' in |$L^y$\verb| is|'.
\end{quote}

\section{Reproducibility Checklist}
\label{appendix:reproduce}
\begin{itemize}[wide, labelwidth=0pt, labelindent=0pt, itemsep=1pt, topsep=1pt]
    \item \textbf{Source Code}: our code is publicly available at \url{https://github.com/cambridgeltl/sail-bli}.
    \item \textbf{Hyper-Parameter Search}: $N_{it}$ is selected from $\{1,2,3,4\}$ and $N_{f}$ from $\{1000, 2000, 3000, 4000,\\5000, 6000, 7000, 8000, 9000, 10000\}$. 
    \item \textbf{Software}: Python $3.9.7$, PyTorch $1.10.1$, Transformers $4.28.1$, OpenAI $1.28.1$.
    \item \textbf{Computing Infrastructure}: we run our codes on \href{https://www.hpc.cam.ac.uk/high-performance-computing}{Wilkes3}, a GPU cluster hosted by the University of Cambridge. Each run makes use of a single Nvidia 80GB A100 GPU and $32\times$ CPU cores.
    
    \item \textbf{Half-Precision Floating-Point Format}: as introduced in \S\ref{s:experiments}, our BLI inference relies on \verb|torch.float16| for both our \sail and the \zero baseline. We have verified that \verb|fp16| can accelerate our computation with only negligible impact on the absolute BLI performance. Note that \citet{li-etal-2023-bilingual} did not specify \verb|torch.float16| in their \zero experiments with \llama$_{\text{7B}}$ and \llama$_{\text{13B}}$, so the BLI scores reported are slightly different from ours.
    \item \textbf{Data, WEs, LLMs}: all the BLI data, WEs, LLMs (excluding \chatgpt models) and baseline codes are open-source and publicly available. The WEs for retrieving in-context examples are fastText WEs~\citep{bojanowski-etal-2017-enriching} trained on monolingual corpora of respective languages: the version pretrained on Wikipedia\footnote{\url{https://fasttext.cc/docs/en/pretrained-vectors.html}} is used for XLING and the version pretrained with Wikipedia plus Common Crawl\footnote{\url{https://fasttext.cc/docs/en/crawl-vectors.html}} is used for PanLex-BLI, as recommended by XLING and PanLex-BLI, respectively. The same WEs are used for our \mapping baselines. The LLMs used in our main experiments (\llama models) are summarised in Table~\ref{table:modelswithsizes}. Note that we only adopt pretrained versions of \llama (e.g., `\texttt{meta-llama/Llama-2-7b-hf}') \textit{rather than} the instruction-tuned models (e.g., `\texttt{meta-llama/Llama-2-7b-chat-hf}'). The details of \chatgpt models used in \S\ref{r:fa} are provided in Appendix~\ref{appendix:chatgpt}.
    \item \textbf{Baselines}: for every baseline, we use its recommended setup for unsupervised BLI and make sure the recommended setup achieves its own (near-)optimal performance. As introduced in \S\ref{s:experiments}, we extend \contrastivebli to the unsupervised BLI setup. Specifically, we adopt the set of its hyper-parameters recommended for the weakly supervised BLI setup, which we found can also achieve strong unsupervised BLI performance. 
    \item \textbf{Parameter Count and Runtime}: we report the number of parameters of each LLM and the GPU runtime for BLI inference on a single BLI direction \textsc{de}$\to$\textsc{fr}, which contains circa $2$K word pairs, in Table~\ref{table:modelswithsizes}. 
    \item \textbf{Carbon Footprint}: our work consumes about $750$ A100 GPU hours in total. We estimate that our experiments causes the emission of circa $90$kg CO$_2$ equivalents according to a publicly available `machine learning emissions calculator'~\cite{luccioni2019quantifying}\footnote{\url{https://mlco2.github.io/impact/\#compute}}.  
\end{itemize}

\begin{table*}[!t]
\begin{center}
\resizebox{0.85\linewidth}{!}{%
\begin{tabular}{lllll}
\toprule 
\rowcolor{Gray}
\multicolumn{1}{c}{\bf LLM} &\multicolumn{1}{c}{\bf Model ID}  &\multicolumn{1}{c}{\bf Parameter Count}&\multicolumn{1}{c}{\bf Runtime: \zero}&\multicolumn{1}{c}{\bf Runtime: \sail}\\
\midrule
\multicolumn{1}{c}{\llama$_{\text{7B}}$} &\multicolumn{1}{c}{`\texttt{huggyllama/llama-7b}'} &\multicolumn{1}{r}{\textsc{$6,738,415,616$}}&\multicolumn{1}{c}{$5$ min}&\multicolumn{1}{c}{$40$ min}\\
\multicolumn{1}{c}{\llama-2$_{\text{7B}}$} &\multicolumn{1}{c}{`\texttt{meta-llama/Llama-2-7b-hf}'} &\multicolumn{1}{r}{\textsc{$6,738,415,616$}}&\multicolumn{1}{c}{$5$ min}&\multicolumn{1}{c}{$40$ min}\\
\multicolumn{1}{c}{\llama$_{\text{13B}}$} &\multicolumn{1}{c}{`\texttt{huggyllama/llama-13b}'} &\multicolumn{1}{r}{\textsc{$13,015,864,320$}}&\multicolumn{1}{c}{$6$ min}&\multicolumn{1}{c}{$49$ min}\\
\multicolumn{1}{c}{\llama-2$_{\text{13B}}$} &\multicolumn{1}{c}{`\texttt{meta-llama/Llama-2-13b-hf}'} &\multicolumn{1}{r}{\textsc{$13,015,864,320$}}&\multicolumn{1}{c}{$6$ min}&\multicolumn{1}{c}{$49$ min}\\
\bottomrule
\end{tabular}
}
\caption{LLMs adopted in our work with their \rurl{huggingface.co} model IDs, parameter count, and GPU runtime on a single BLI direction for \zero prompting and \sail respectively.}
\label{table:modelswithsizes}
\end{center}
\end{table*}

\section{Details of \chatgpt Experiments}
\label{appendix:chatgpt}

We run our \chatgpt experiments introduced in \S\ref{r:fa} with the OpenAI API.\footnote{\url{https://platform.openai.com/docs/overview}} The model ID for \gpt-3.5 is `\texttt{gpt-3.5-turbo-0125}'. For \gpt-4, we adopt the state-of-the-art `\texttt{gpt-4-turbo-2024-04-09}' model which ranked $1^{\text{st}}$ on the \href{https://chat.lmsys.org/?leaderboard}{LMSYS Chatbot Arena Leaderboard} at the time of experimentation (May 12, 2024).

Our input to \chatgpt consists of two types of input messages: a \textit{system} message followed by a \textit{user} message. For the user message, we adopt the following template for both \gpt-3.5 and \gpt-4 as recommended in~\citet{dmbli}:
\begin{quote}
`\verb|Translate the |$L^x$\verb| word |$w^{x}$\verb| into |$L^y$\verb|:|',
\end{quote}
which is also selected from the template pool of~\citet{li-etal-2023-bilingual}. We additionally adopt the following system message which is not used in~\citet{dmbli} or~\citet{li-etal-2023-bilingual}:
\begin{quote}
`\verb|Please complete the following|\\\verb|sentence and only output the target|\\\verb|word.|'.
\end{quote}
In our preliminary investigation, we find that our system message can considerably improve the BLI performance of both \chatgpt models.

There are two hyper-parameters used in our API calls: $\verb|temperature|=0$ and $\verb|max_tokens|=5$. Like our main experiments, we also extract the first word in the generated output sequence as the prediction for the target word. But different from our \llama experiments, we only derive a single output sequence from the \chatgpt API for each prompt. The code for our \chatgpt experiments is also provided in our GitHub repository.

\section{Full BLI Results}
\label{appendix:res}

Table~\ref{table:mainresappendix} shows detailed BLI scores for each BLI direction in the XLING dataset. Similarly, individual per-direction results on PanLex-BLI are presented in Table~\ref{table:mainrespappendix}.

\begin{table*}[ht]
\begin{center}
\resizebox{1.0\textwidth}{!}{%
\begin{tabular}{llllllllllll}
\toprule 
\rowcolor{Gray}
\multicolumn{1}{c}{\bf[Unsupervised BLI]} &\multicolumn{1}{c}{\bf\vecmap} &\multicolumn{1}{c}{\bf\contrastivebli (C1)} &\multicolumn{1}{c}{\bf\contrastivebli (C2)} &\multicolumn{1}{c}{\bf\llama$_{\text{7B}}$} &\multicolumn{1}{c}{\bf\llama-2$_{\text{7B}}$} &\multicolumn{1}{c}{\bf\llama$_{\text{13B}}$} &\multicolumn{1}{c}{\bf\llama-2$_{\text{13B}}$} &\multicolumn{1}{c}{\bf\llama$_{\text{7B}}$} &\multicolumn{1}{c}{\bf\llama-2$_{\text{7B}}$} &\multicolumn{1}{c}{\bf\llama$_{\text{13B}}$} &\multicolumn{1}{c}{\bf\llama-2$_{\text{13B}}$} \\

\cmidrule(lr){2-4} \cmidrule(lr){5-8} \cmidrule(lr){9-12}

\multicolumn{1}{c}{} &\multicolumn{3}{c}{\bf\mapping} &\multicolumn{4}{c}{\bf\zero} &\multicolumn{4}{c}{\bf\sail (Ours)}\\
 
\multicolumn{1}{c}{\textsc{de}$\to$\textsc{fr}} &\multicolumn{1}{c}{48.98} &\multicolumn{1}{c}{50.39} &\multicolumn{1}{c}{51.8} &\multicolumn{1}{c}{42.46} &\multicolumn{1}{c}{44.44} &\multicolumn{1}{c}{47.37} &\multicolumn{1}{c}{46.64} &\multicolumn{1}{c}{54.67} &\multicolumn{1}{c}{54.77} &\multicolumn{1}{c}{58.37} &\multicolumn{1}{c}{\bf61.5}\\
\multicolumn{1}{c}{\textsc{fr}$\to$\textsc{de}} &\multicolumn{1}{c}{43.97} &\multicolumn{1}{c}{43.61} &\multicolumn{1}{c}{44.9} &\multicolumn{1}{c}{43.2} &\multicolumn{1}{c}{45.47} &\multicolumn{1}{c}{48.11} &\multicolumn{1}{c}{50.8} &\multicolumn{1}{c}{50.08} &\multicolumn{1}{c}{54.16} &\multicolumn{1}{c}{54.47} &\multicolumn{1}{c}{\bf56.29}\\
\multicolumn{1}{c}{\textsc{de}$\to$\textsc{it}} &\multicolumn{1}{c}{48.41} &\multicolumn{1}{c}{49.77} &\multicolumn{1}{c}{50.23} &\multicolumn{1}{c}{42.78} &\multicolumn{1}{c}{42.78} &\multicolumn{1}{c}{46.06} &\multicolumn{1}{c}{48.51} &\multicolumn{1}{c}{53.36} &\multicolumn{1}{c}{54.25} &\multicolumn{1}{c}{57.38} &\multicolumn{1}{c}{\bf59.05}\\
\multicolumn{1}{c}{\textsc{it}$\to$\textsc{de}} &\multicolumn{1}{c}{44.03} &\multicolumn{1}{c}{43.93} &\multicolumn{1}{c}{45.43} &\multicolumn{1}{c}{38.6} &\multicolumn{1}{c}{41.55} &\multicolumn{1}{c}{44.39} &\multicolumn{1}{c}{45.27} &\multicolumn{1}{c}{46.15} &\multicolumn{1}{c}{51.63} &\multicolumn{1}{c}{52.2} &\multicolumn{1}{c}{\bf52.92}\\
\multicolumn{1}{c}{\textsc{de}$\to$\textsc{ru}} &\multicolumn{1}{c}{25.67} &\multicolumn{1}{c}{28.22} &\multicolumn{1}{c}{31.09} &\multicolumn{1}{c}{30.41} &\multicolumn{1}{c}{35.32} &\multicolumn{1}{c}{32.76} &\multicolumn{1}{c}{36.62} &\multicolumn{1}{c}{45.12} &\multicolumn{1}{c}{46.9} &\multicolumn{1}{c}{48.98} &\multicolumn{1}{c}{\bf51.59}\\
\multicolumn{1}{c}{\textsc{ru}$\to$\textsc{de}} &\multicolumn{1}{c}{39.13} &\multicolumn{1}{c}{40.02} &\multicolumn{1}{c}{41.33} &\multicolumn{1}{c}{43.53} &\multicolumn{1}{c}{44.68} &\multicolumn{1}{c}{43.11} &\multicolumn{1}{c}{42.12} &\multicolumn{1}{c}{46.83} &\multicolumn{1}{c}{50.55} &\multicolumn{1}{c}{50.65} &\multicolumn{1}{c}{\bf53.9}\\
\multicolumn{1}{c}{\textsc{en}$\to$\textsc{de}} &\multicolumn{1}{c}{48.4} &\multicolumn{1}{c}{47.45} &\multicolumn{1}{c}{47.4} &\multicolumn{1}{c}{52.0} &\multicolumn{1}{c}{52.1} &\multicolumn{1}{c}{54.35} &\multicolumn{1}{c}{59.85} &\multicolumn{1}{c}{59.55} &\multicolumn{1}{c}{61.75} &\multicolumn{1}{c}{62.8} &\multicolumn{1}{c}{\bf65.05}\\
\multicolumn{1}{c}{\textsc{de}$\to$\textsc{en}} &\multicolumn{1}{c}{54.51} &\multicolumn{1}{c}{54.36} &\multicolumn{1}{c}{55.97} &\multicolumn{1}{c}{42.57} &\multicolumn{1}{c}{44.91} &\multicolumn{1}{c}{46.95} &\multicolumn{1}{c}{47.16} &\multicolumn{1}{c}{55.35} &\multicolumn{1}{c}{56.44} &\multicolumn{1}{c}{57.96} &\multicolumn{1}{c}{\bf61.24}\\
\multicolumn{1}{c}{\textsc{en}$\to$\textsc{fr}} &\multicolumn{1}{c}{60.15} &\multicolumn{1}{c}{61.05} &\multicolumn{1}{c}{61.25} &\multicolumn{1}{c}{57.6} &\multicolumn{1}{c}{62.65} &\multicolumn{1}{c}{62.65} &\multicolumn{1}{c}{61.75} &\multicolumn{1}{c}{72.6} &\multicolumn{1}{c}{73.8} &\multicolumn{1}{c}{75.85} &\multicolumn{1}{c}{\bf76.35}\\
\multicolumn{1}{c}{\textsc{fr}$\to$\textsc{en}} &\multicolumn{1}{c}{61.25} &\multicolumn{1}{c}{62.34} &\multicolumn{1}{c}{63.58} &\multicolumn{1}{c}{54.58} &\multicolumn{1}{c}{55.56} &\multicolumn{1}{c}{57.27} &\multicolumn{1}{c}{53.03} &\multicolumn{1}{c}{63.68} &\multicolumn{1}{c}{65.13} &\multicolumn{1}{c}{65.29} &\multicolumn{1}{c}{\bf66.63}\\
\multicolumn{1}{c}{\textsc{en}$\to$\textsc{it}} &\multicolumn{1}{c}{57.4} &\multicolumn{1}{c}{57.6} &\multicolumn{1}{c}{58.75} &\multicolumn{1}{c}{58.95} &\multicolumn{1}{c}{60.85} &\multicolumn{1}{c}{60.4} &\multicolumn{1}{c}{65.8} &\multicolumn{1}{c}{71.7} &\multicolumn{1}{c}{73.0} &\multicolumn{1}{c}{74.25} &\multicolumn{1}{c}{\bf77.6}\\
\multicolumn{1}{c}{\textsc{it}$\to$\textsc{en}} &\multicolumn{1}{c}{60.83} &\multicolumn{1}{c}{62.02} &\multicolumn{1}{c}{63.46} &\multicolumn{1}{c}{47.39} &\multicolumn{1}{c}{50.08} &\multicolumn{1}{c}{54.94} &\multicolumn{1}{c}{53.54} &\multicolumn{1}{c}{60.1} &\multicolumn{1}{c}{64.08} &\multicolumn{1}{c}{64.13} &\multicolumn{1}{c}{\bf65.43}\\
\multicolumn{1}{c}{\textsc{en}$\to$\textsc{ru}} &\multicolumn{1}{c}{24.55} &\multicolumn{1}{c}{25.45} &\multicolumn{1}{c}{26.1} &\multicolumn{1}{c}{42.05} &\multicolumn{1}{c}{44.6} &\multicolumn{1}{c}{40.1} &\multicolumn{1}{c}{47.6} &\multicolumn{1}{c}{57.4} &\multicolumn{1}{c}{60.25} &\multicolumn{1}{c}{61.05} &\multicolumn{1}{c}{\bf63.75}\\
\multicolumn{1}{c}{\textsc{ru}$\to$\textsc{en}} &\multicolumn{1}{c}{46.52} &\multicolumn{1}{c}{46.67} &\multicolumn{1}{c}{50.03} &\multicolumn{1}{c}{46.15} &\multicolumn{1}{c}{50.81} &\multicolumn{1}{c}{50.13} &\multicolumn{1}{c}{51.44} &\multicolumn{1}{c}{54.95} &\multicolumn{1}{c}{58.51} &\multicolumn{1}{c}{57.41} &\multicolumn{1}{c}{\bf59.93}\\
\multicolumn{1}{c}{\textsc{it}$\to$\textsc{fr}} &\multicolumn{1}{c}{64.75} &\multicolumn{1}{c}{65.12} &\multicolumn{1}{c}{65.89} &\multicolumn{1}{c}{51.42} &\multicolumn{1}{c}{54.47} &\multicolumn{1}{c}{57.36} &\multicolumn{1}{c}{55.3} &\multicolumn{1}{c}{61.91} &\multicolumn{1}{c}{65.58} &\multicolumn{1}{c}{65.94} &\multicolumn{1}{c}{\bf68.17}\\
\multicolumn{1}{c}{\textsc{fr}$\to$\textsc{it}} &\multicolumn{1}{c}{63.37} &\multicolumn{1}{c}{63.94} &\multicolumn{1}{c}{64.61} &\multicolumn{1}{c}{57.32} &\multicolumn{1}{c}{55.98} &\multicolumn{1}{c}{60.01} &\multicolumn{1}{c}{61.87} &\multicolumn{1}{c}{64.72} &\multicolumn{1}{c}{66.22} &\multicolumn{1}{c}{69.22} &\multicolumn{1}{c}{\bf69.53}\\
\multicolumn{1}{c}{\textsc{ru}$\to$\textsc{fr}} &\multicolumn{1}{c}{45.31} &\multicolumn{1}{c}{46.78} &\multicolumn{1}{c}{47.93} &\multicolumn{1}{c}{43.58} &\multicolumn{1}{c}{48.04} &\multicolumn{1}{c}{47.77} &\multicolumn{1}{c}{41.17} &\multicolumn{1}{c}{54.79} &\multicolumn{1}{c}{57.62} &\multicolumn{1}{c}{57.52} &\multicolumn{1}{c}{\bf60.29}\\
\multicolumn{1}{c}{\textsc{fr}$\to$\textsc{ru}} &\multicolumn{1}{c}{24.26} &\multicolumn{1}{c}{25.09} &\multicolumn{1}{c}{26.07} &\multicolumn{1}{c}{35.8} &\multicolumn{1}{c}{38.8} &\multicolumn{1}{c}{38.59} &\multicolumn{1}{c}{39.94} &\multicolumn{1}{c}{48.94} &\multicolumn{1}{c}{51.42} &\multicolumn{1}{c}{53.29} &\multicolumn{1}{c}{\bf54.11}\\
\multicolumn{1}{c}{\textsc{ru}$\to$\textsc{it}} &\multicolumn{1}{c}{43.95} &\multicolumn{1}{c}{44.89} &\multicolumn{1}{c}{46.15} &\multicolumn{1}{c}{47.3} &\multicolumn{1}{c}{47.15} &\multicolumn{1}{c}{45.99} &\multicolumn{1}{c}{49.45} &\multicolumn{1}{c}{53.54} &\multicolumn{1}{c}{56.26} &\multicolumn{1}{c}{56.31} &\multicolumn{1}{c}{\bf59.25}\\
\multicolumn{1}{c}{\textsc{it}$\to$\textsc{ru}} &\multicolumn{1}{c}{25.48} &\multicolumn{1}{c}{26.87} &\multicolumn{1}{c}{29.35} &\multicolumn{1}{c}{31.52} &\multicolumn{1}{c}{33.02} &\multicolumn{1}{c}{35.45} &\multicolumn{1}{c}{36.38} &\multicolumn{1}{c}{44.03} &\multicolumn{1}{c}{48.63} &\multicolumn{1}{c}{50.75} &\multicolumn{1}{c}{\bf53.49}\\
\multicolumn{1}{c}{Avg.} &\multicolumn{1}{c}{46.55} &\multicolumn{1}{c}{47.28} &\multicolumn{1}{c}{48.57} &\multicolumn{1}{c}{45.46} &\multicolumn{1}{c}{47.66} &\multicolumn{1}{c}{48.69} &\multicolumn{1}{c}{49.71} &\multicolumn{1}{c}{55.97} &\multicolumn{1}{c}{58.55} &\multicolumn{1}{c}{59.69} &\multicolumn{1}{c}{\bf61.8}\\
\bottomrule
\end{tabular}
}
\caption{Full BLI results on $20$ XLING BLI directions.}
\label{table:mainresappendix}
\end{center}
\end{table*}

\begin{table*}[ht]
\begin{center}
\resizebox{1.0\textwidth}{!}{%
\begin{tabular}{llllllllllll}
\toprule 
\rowcolor{Gray}
\multicolumn{1}{c}{\bf[Unsupervised BLI]} &\multicolumn{1}{c}{\bf\vecmap} &\multicolumn{1}{c}{\bf\contrastivebli (C1)} &\multicolumn{1}{c}{\bf\contrastivebli (C2)} &\multicolumn{1}{c}{\bf\llama$_{\text{7B}}$} &\multicolumn{1}{c}{\bf\llama-2$_{\text{7B}}$} &\multicolumn{1}{c}{\bf\llama$_{\text{13B}}$} &\multicolumn{1}{c}{\bf\llama-2$_{\text{13B}}$} &\multicolumn{1}{c}{\bf\llama$_{\text{7B}}$} &\multicolumn{1}{c}{\bf\llama-2$_{\text{7B}}$} &\multicolumn{1}{c}{\bf\llama$_{\text{13B}}$} &\multicolumn{1}{c}{\bf\llama-2$_{\text{13B}}$} \\

\cmidrule(lr){2-4} \cmidrule(lr){5-8} \cmidrule(lr){9-12}

\multicolumn{1}{c}{} &\multicolumn{3}{c}{\bf\mapping} &\multicolumn{4}{c}{\bf\zero} &\multicolumn{4}{c}{\bf\sail (Ours)}\\
\multicolumn{1}{c}{\textsc{bg}$\to$\textsc{ca}} &\multicolumn{1}{c}{39.6} &\multicolumn{1}{c}{38.08} &\multicolumn{1}{c}{39.66} &\multicolumn{1}{c}{32.83} &\multicolumn{1}{c}{29.79} &\multicolumn{1}{c}{32.77} &\multicolumn{1}{c}{33.47} &\multicolumn{1}{c}{40.19} &\multicolumn{1}{c}{42.23} &\multicolumn{1}{c}{42.52} &\multicolumn{1}{c}{\bf47.9}\\
\multicolumn{1}{c}{\textsc{ca}$\to$\textsc{hu}} &\multicolumn{1}{c}{34.09} &\multicolumn{1}{c}{34.2} &\multicolumn{1}{c}{36.85} &\multicolumn{1}{c}{23.7} &\multicolumn{1}{c}{23.2} &\multicolumn{1}{c}{24.42} &\multicolumn{1}{c}{30.17} &\multicolumn{1}{c}{32.27} &\multicolumn{1}{c}{35.25} &\multicolumn{1}{c}{38.34} &\multicolumn{1}{c}{\bf39.83}\\
\multicolumn{1}{c}{\textsc{hu}$\to$\textsc{bg}} &\multicolumn{1}{c}{36.46} &\multicolumn{1}{c}{38.36} &\multicolumn{1}{c}{40.44} &\multicolumn{1}{c}{28.28} &\multicolumn{1}{c}{27.71} &\multicolumn{1}{c}{26.5} &\multicolumn{1}{c}{26.73} &\multicolumn{1}{c}{38.19} &\multicolumn{1}{c}{41.47} &\multicolumn{1}{c}{43.89} &\multicolumn{1}{c}{\bf46.66}\\
\multicolumn{1}{c}{\textsc{ca}$\to$\textsc{bg}} &\multicolumn{1}{c}{33.6} &\multicolumn{1}{c}{31.39} &\multicolumn{1}{c}{33.94} &\multicolumn{1}{c}{26.35} &\multicolumn{1}{c}{27.2} &\multicolumn{1}{c}{27.03} &\multicolumn{1}{c}{28.39} &\multicolumn{1}{c}{36.54} &\multicolumn{1}{c}{38.47} &\multicolumn{1}{c}{42.27} &\multicolumn{1}{c}{\bf45.67}\\
\multicolumn{1}{c}{\textsc{hu}$\to$\textsc{ca}} &\multicolumn{1}{c}{37.79} &\multicolumn{1}{c}{39.77} &\multicolumn{1}{c}{43.45} &\multicolumn{1}{c}{32.62} &\multicolumn{1}{c}{28.66} &\multicolumn{1}{c}{38.23} &\multicolumn{1}{c}{37.51} &\multicolumn{1}{c}{41.53} &\multicolumn{1}{c}{46.09} &\multicolumn{1}{c}{47.91} &\multicolumn{1}{c}{\bf51.65}\\
\multicolumn{1}{c}{\textsc{bg}$\to$\textsc{hu}} &\multicolumn{1}{c}{39.24} &\multicolumn{1}{c}{38.95} &\multicolumn{1}{c}{\bf41.44} &\multicolumn{1}{c}{24.13} &\multicolumn{1}{c}{28.12} &\multicolumn{1}{c}{23.67} &\multicolumn{1}{c}{27.72} &\multicolumn{1}{c}{33.16} &\multicolumn{1}{c}{38.08} &\multicolumn{1}{c}{38.14} &\multicolumn{1}{c}{41.38}\\
\multicolumn{1}{c}{Avg.} &\multicolumn{1}{c}{36.8} &\multicolumn{1}{c}{36.79} &\multicolumn{1}{c}{39.3} &\multicolumn{1}{c}{27.98} &\multicolumn{1}{c}{27.45} &\multicolumn{1}{c}{28.77} &\multicolumn{1}{c}{30.66} &\multicolumn{1}{c}{36.98} &\multicolumn{1}{c}{40.27} &\multicolumn{1}{c}{42.18} &\multicolumn{1}{c}{\bf45.51}\\
\bottomrule
\end{tabular}
}
\caption{Full BLI results on $6$ PanLex-BLI BLI directions.}
\label{table:mainrespappendix}
\end{center}
\end{table*}

\end{document}